\documentclass[11pt, a4paper, logo, copyright, nonumbering]{deepseek}
\usepackage[authoryear, sort&compress, round]{natbib}
\usepackage{dblfloatfix}
\usepackage{ulem}
\usepackage{caption}
\usepackage{subcaption}
\usepackage{dramatist}
\usepackage{xspace}
\usepackage{pifont}
\usepackage{multirow}
\usepackage{tcolorbox}
\usepackage{xltabular}
\usepackage{longtable}
\usepackage{hyperref}
\usepackage{amsfonts}
\usepackage{amsmath}
\usepackage{amssymb}
\usepackage{lineno}
\usepackage{cleveref}
\usepackage{makecell}

\usepackage[bottom]{footmisc}

\usepackage{CJKutf8}
\usepackage{setspace}

\usepackage{dsfont}
\usepackage[table]{xcolor}
\usepackage{tabularx}
\usepackage{booktabs}
\newcolumntype{Y}{>{\centering\arraybackslash}X}

\newcommand{\hpre}[1]{\ensuremath{\mathcal{H}^{\mathrm{pre}}_{#1}}}
\newcommand{\hpost}[1]{\ensuremath{\mathcal{H}^{\mathrm{post}}_{#1}}}
\newcommand{\hres}[1]{\ensuremath{\mathcal{H}^{\mathrm{res}}_{#1}}}
\newcommand{\tlhpre}[1]{\ensuremath{\tilde{\mathcal{H}}^{\mathrm{pre}}_{#1}}}
\newcommand{\tlhpost}[1]{\ensuremath{\tilde{\mathcal{H}}^{\mathrm{post}}_{#1}}}
\newcommand{\tlhres}[1]{\ensuremath{\tilde{\mathcal{H}}^{\mathrm{res}}_{#1}}}
\newcommand{\ttlhpre}[1]{\ensuremath{\tilde{\tilde{\mathcal{H}}}^{\mathrm{pre}}_{#1}}}
\newcommand{\ttlhpost}[1]{\ensuremath{\tilde{\tilde{\mathcal{H}}}^{\mathrm{post}}_{#1}}}
\newcommand{\ttlhres}[1]{\ensuremath{\tilde{\tilde{\mathcal{H}}}^{\mathrm{res}}_{#1}}}

\makeatletter
\def\@BTrule[#1]{%
  \ifx\longtable\undefined
    \let\@BTswitch\@BTnormal
  \else\ifx\hline\LT@hline
    \nobreak
    \let\@BTswitch\@BLTrule
  \else
     \let\@BTswitch\@BTnormal
  \fi\fi
  \global\@thisrulewidth=#1\relax
  \ifnum\@thisruleclass=\tw@\vskip\@aboverulesep\else
  \ifnum\@lastruleclass=\z@\vskip\@aboverulesep\else
  \ifnum\@lastruleclass=\@ne\vskip\doublerulesep\fi\fi\fi
  \@BTswitch}
\makeatother

\addto\extrasenglish{
}

 {\begin{list}{}%
         {\setlength{\leftmargin}{#1}}%
         \item[]%
 }
 {\end{list}}
 
\reportnumber{001} 

\renewcommand{\today}{}

\title{
\vspace{-0.2in}
\centering \textit{m}HC: Manifold-Constrained Hyper-Connections \vspace{-0.2in}
}

\author[]{
Zhenda Xie*$^\dag$, Yixuan Wei*, Huanqi Cao*,

\vspace{-0.1in}
Chenggang Zhao, Chengqi Deng, Jiashi Li, Damai Dai, Huazuo Gao, Jiang Chang,

\vspace{-0.1in}
Kuai Yu, Liang Zhao, Shangyan Zhou, Zhean Xu, Zhengyan Zhang, Wangding Zeng,

\vspace{-0.1in}
Shengding Hu, Yuqing Wang, Jingyang Yuan, Lean Wang, Wenfeng Liang \\

DeepSeek-AI

\vspace{-0.2in}
}

\correspondingauthor{Core contributors. $^\dag$Corresponding author: \href{mailto:xie.zhenda@deepseek.com}{xie.zhenda@deepseek.com}}

\renewcommand{\phi}{\varphi}

\renewcommand{\geq}{\geqslant}

\renewcommand{\epsilon}{\varepsilon}
\renewcommand{\imath}{\mathrm{i}}

\newlength{\restsubwidth}
\newlength{\restsubheight}
\newlength{\restsubmoreheight}
\newcommand{\rest}[2]{%
        \settowidth{\restsubwidth}{\ensuremath{#2}}
        \settoheight{\restsubheight}{\ensuremath{{}_{#2}}}
        \ensuremath{{#1\hskip 0.5pt}_{\vrule\kern2pt\parbox[b][%
        4pt][b]{\the\restsubwidth}{%
                        \ensuremath{{}_{#2}}}}}
        }

\newcommand{\mhcshort}{\textit{m}HC}
\newcommand{\mhcfull}{Manifold-Constrained Hyper-Connections}

\newcommand{\mhcshortbf}{\textbf{\textit{m}HC}}
\newcommand{\mhcfullbf}{\textbf{Manifold-Constrained Hyper-Connections}}

\begin{abstract}

\vspace{-0.2in}

Recently, studies exemplified by Hyper-Connections (HC) have extended the ubiquitous residual connection paradigm established over the past decade by expanding the residual stream width and diversifying connectivity patterns.
While yielding substantial performance gains, this diversification fundamentally compromises the identity mapping property intrinsic to the residual connection, which causes severe training instability and restricted scalability, and additionally incurs notable memory access overhead.
To address these challenges, we propose \mhcfullbf{} (\mhcshortbf{}), a general framework that projects the residual connection space of HC onto a specific manifold to restore the identity mapping property, while incorporating rigorous infrastructure optimization to ensure efficiency.
Empirical experiments demonstrate that \mhcshort{} is effective for training at scale, offering tangible performance improvements and superior scalability.
We anticipate that \mhcshort{}, as a flexible and practical extension of HC, will contribute to a deeper understanding of topological architecture design and suggest promising directions for the evolution of foundational models.

\end{abstract}

\begin{document}
\begin{CJK*}{UTF8}{gbsn}

\maketitle

\begin{figure}[h]
    \centering
    \includegraphics[width=1.0\textwidth]{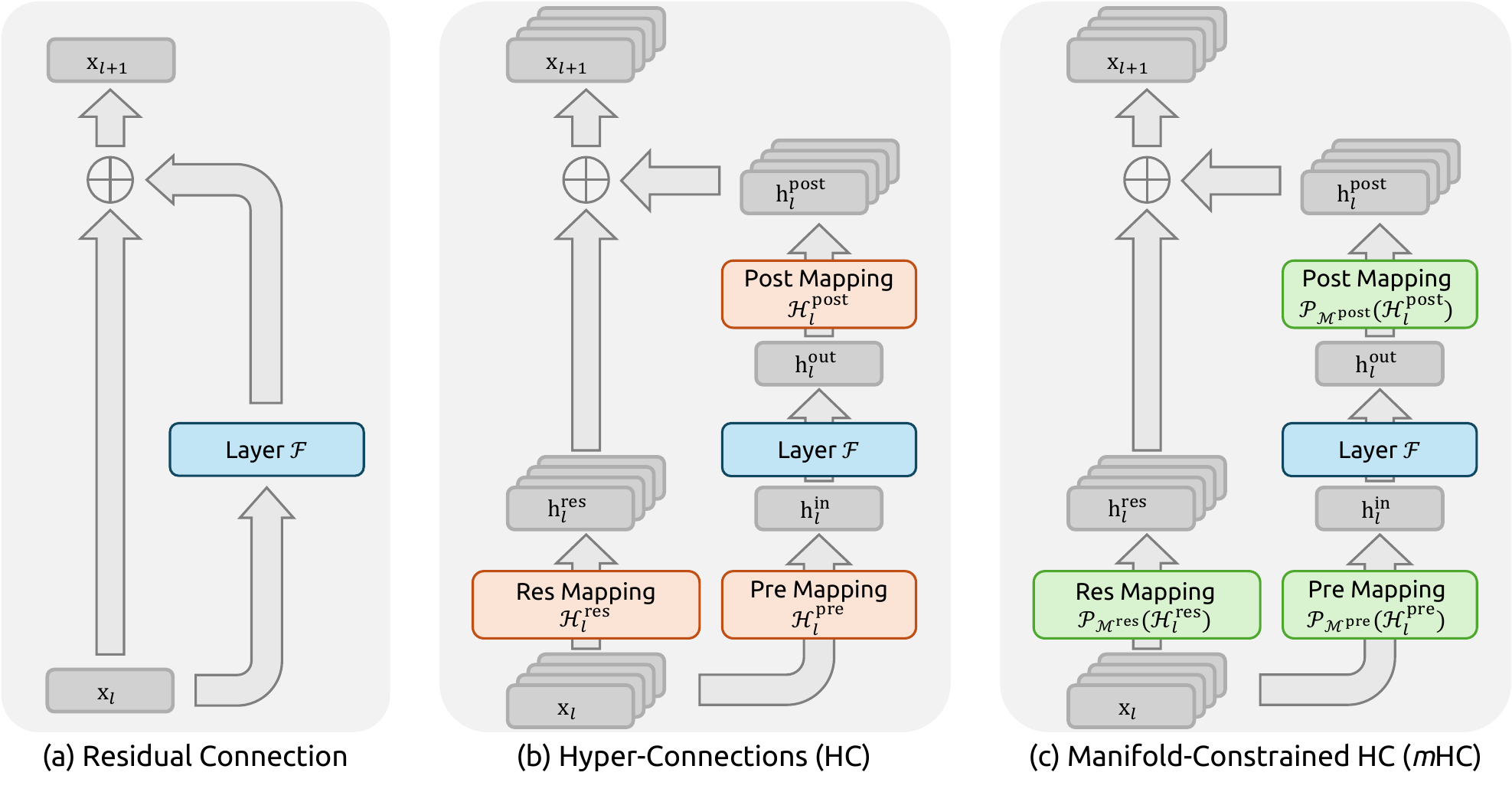}
    \vspace{-0.2in}
    \caption{
    \textbf{Illustrations of Residual Connection Paradigms.} This figure compares the structural design of (a) standard Residual Connection, (b) Hyper-Connections (HC), and (c) our proposed \mhcfullbf{} (\mhcshortbf{}). Unlike the unconstrained HC, \mhcshort{} focuses on optimizing the residual connection space by projecting the matrices onto a constrained manifold to ensure stability.
    }
    \label{fig:teaser}
    \vspace{-0.2in}
\end{figure}

\newpage

\begin{spacing}{0.9}
\tableofcontents
\end{spacing}

\newpage

\section{Introduction}

Deep neural network architectures have undergone rapid evolution since the introduction of ResNets~\citep{he2016deep}.
As illustrated in Fig.~\ref{fig:teaser}(a), the structure of a single-layer can be formulated as follows:
\begin{equation}
    \mathbf{x}_{l+1} = \mathbf{x}_l + \mathcal{F}(\mathbf{x}_l, \mathcal{W}_l),
    \label{eqn:single_rc}
\end{equation}
where $\mathbf{x}_l$ and $\mathbf{x}_{l+1}$ denote the $C$-dimensional input and output of the $l$-th layer, respectively, and $\mathcal{F}$ represents the residual function.
Although the residual function $\mathcal{F}$ has evolved over the past decade to include various operations such as convolution, attention mechanisms, and feed forward networks, the paradigm of the residual connection has maintained its original form. Accompanying the progression of Transformer~\citep{vaswani2017attention} architecture, this paradigm has currently established itself as a fundamental design element in large language models~(LLMs)~\citep{brown2020language, touvron2023llama, liu2024deepseek_v3}.

This success is primarily attributed to the concise form of the residual connection. More importantly, early research~\citep{he2016identity} revealed that the identity mapping property of the residual connection maintains stability and efficiency during large-scale training. By recursively extending the residual connection across multiple layers, Eq.~\eqref{eqn:single_rc} yields:
\begin{equation}
    \mathbf{x}_L = \mathbf{x}_l + \sum_{i=l}^{L-1} \mathcal{F}(\mathbf{x}_i, \mathcal{W}_i),
    \label{eqn:multi_rc}
\end{equation}
where $L$ and $l$ correspond to deeper and shallower layers, respectively.
The term identity mapping refers to the component $\mathbf{x}_l$ itself, which emphasizes the property that the signal from the shallower layer maps directly to the deeper layer without any modification.

Recently, studies exemplified by Hyper-Connections (HC)~\citep{zhu2024hyper} have introduced a new dimension to the residual connection and empirically demonstrated its performance potential. The single-layer architecture of HC is illustrated in Fig.~\ref{fig:teaser}(b).
By expanding the width of the residual stream and enhancing connection complexity, HC significantly increases topological complexity without altering the computational overhead of individual units regarding FLOPs.
Formally, single-layer propagation in HC is defined as:
\begin{equation}
    \mathbf{x}_{l+1} = \mathcal{H}_{l}^{\mathrm{res}}\mathbf{x}_l + \mathcal{H}_{l}^{\mathrm{post}\, \top}\mathcal{F}(\mathcal{H}_{l}^{\mathrm{pre}}\mathbf{x}_l, \mathcal{W}_l),
    \label{eqn:single_hc}
\end{equation}
where $\mathbf{x}_{l}$ and $\mathbf{x}_{l+1}$ denote the input and output of the $l$-th layer, respectively.
Unlike the formulation in Eq.~\eqref{eqn:single_rc}, the feature dimension of $\mathbf{x}_{l}$ and $\mathbf{x}_{l+1}$ is expanded from $C$ to $n \times C$, where $n$ is the expansion rate.
The term $\mathcal{H}_{l}^{\mathrm{res}} \in \mathbb{R}^{n \times n}$ represents a learnable mapping that mixes features within the residual stream.
Also as a learnable mapping, $\mathcal{H}_{l}^{\mathrm{pre}} \in \mathbb{R}^{1 \times n}$ aggregates features from the $nC$-dim stream into a $C$-dim layer input, and conversely, $\mathcal{H}_{l}^{\mathrm{post}} \in \mathbb{R}^{1 \times n}$ maps the layer output back onto the stream.

However, as the training scale increases, HC introduces potential risks of instability.
The primary concern is that the unconstrained nature of HC compromises the identity mapping property when the architecture extends across multiple layers.
In architectures comprising multiple parallel streams, an ideal identity mapping serves as a conservation mechanism. It ensures that the average signal intensity across streams remains invariant during both forward and backward propagation.
Recursively extending HC to multiple layers via Eq.~\eqref{eqn:single_hc} yields:
\begin{equation}
    \mathbf{x}_{L} = \left(\prod_{i=1}^{L-l}\mathcal{H}_{L-i}^{\mathrm{res}}\right)\mathbf{x}_l + \sum_{i=l}^{L-1}\left(\prod_{j=1}^{L-1-i}\mathcal{H}_{L-j}^{\mathrm{res}}\right)\mathcal{H}_{i}^{\mathrm{post}\, \top}\mathcal{F}(\mathcal{H}_{i}^{\mathrm{pre}}\mathbf{x}_i, \mathcal{W}_i),
    \label{eqn:multi_hc}
\end{equation}
where $L$ and $l$ represent a deeper layer and a shallower layer, respectively.
In contrast to Eq.~\eqref{eqn:multi_rc}, the composite mapping $\prod_{i=1}^{L-l}\mathcal{H}_{L-i}^{\mathrm{res}}$ in HC fails to preserve the global mean of the features. This discrepancy leads to unbounded signal amplification or attenuation, resulting in instability during large-scale training.
A further consideration is that, while HC preserves computational efficiency in terms of FLOPs, the hardware efficiency concerning memory access costs for the widened residual stream remains unaddressed in the original design. These factors collectively restrict the practical scalability of HC and hinder its application in large-scale training.

To address these challenges, we propose \mhcfullbf{} (\mhcshortbf{}), as shown in Fig.~\ref{fig:teaser}(c), a general framework that projects the residual connection space of HC onto a specific manifold to restore the identity mapping property, while incorporating rigorous infrastructure optimization to ensure efficiency.
Specifically, \mhcshort{} utilizes the Sinkhorn-Knopp algorithm~\citep{sinkhorn1967concerning} to entropically project $\mathcal{H}_{l}^{\mathrm{res}}$ onto the Birkhoff polytope. This operation effectively constrains the residual connection matrices within the manifold that is constituted by doubly stochastic matrices.
Since the row and column sums of these matrices equal to $1$, the operation $\mathcal{H}_{l}^{\mathrm{res}}\mathbf{x}_l$ functions as a convex combination of the input features. This characteristic facilitates a well-conditioned signal propagation where the feature mean is conserved, and the signal norm is strictly regularized, effectively mitigating the risk of vanishing or exploding signals. Furthermore, due to the closure of matrix multiplication for doubly stochastic matrices, the composite mapping $\prod_{i=1}^{L-l}\mathcal{H}_{L-i}^{\mathrm{res}}$ retains this conservation property. Consequently, \mhcshort{} effectively maintains the stability of identity mappings between arbitrary depths. To ensure efficiency, we employ kernel fusion and develop mixed precision kernels utilizing TileLang~\citep{wang2025tilelang}. Furthermore, we mitigate the memory footprint through selective recomputing and carefully overlap communication within the DualPipe schedule~\citep{liu2024deepseek_v3}.

Extensive experiments on language model pretraining demonstrate that \mhcshort{} exhibits exceptional stability and scalability while maintaining the performance advantages of HC. In-house large-scale training indicates that \mhcshort{} supports training at scale and introduces only a 6.7\% additional time overhead when expansion rate $n=4$.

\section{Related Works}

Architectural advancements in deep learning can be primarily classified into \textit{micro-design} and \textit{macro-design}.
Micro-design concerns the internal architecture of computational blocks, specifying how features are processed across spatial, temporal, and channel dimensions.
In contrast, macro-design establishes the inter-block topological structure, thereby dictating how feature representations are propagated, routed, and merged across distinct layers.

\subsection{Micro Design}

Driven by parameter sharing and translation invariance, convolution initially dominated the processing of structured signals. While subsequent variations such as depthwise separable~\citep{chollet2017xception} and grouped convolutions~\citep{xie2017aggregated} optimized efficiency, the advent of Transformers~\citep{vaswani2017attention} established Attention and Feed-Forward Networks (FFNs) as the fundamental building blocks of modern architecture. Attention mechanisms facilitate global information propagation, while FFNs enhance the representational capacity of individual features. To balance performance with the computational demands of LLMs, attention mechanisms have evolved towards efficient variants such as Multi-Query Attention (MQA)~\citep{shazeer2019fast}, Grouped-Query Attention (GQA)~\citep{ainslie2023gqa}, and Multi-Head Latent Attention (MLA)~\citep{liu2024deepseek}. Simultaneously, FFNs have been generalized into sparse computing paradigms via Mixture-of-Experts (MoE)~\citep{shazeer2017outrageously, lepikhin2020gshard, fedus2022switch}, allowing for massive parameter scaling without proportional computational costs.

\subsection{Macro Design}

Macro-design governs the global topology of the network~\citep{Srivastava2015highway}. Following ResNet~\citep{he2016deep}, architectures such as DenseNet~\citep{huang2017densely} and FractalNet~\citep{larsson2016fractalnet} aimed to enhance performance by increasing topological complexity through dense connectivity and multi-path structures, respectively. Deep Layer Aggregation (DLA)~\citep{yu2018deep} further extended this paradigm by recursively aggregating features across various depths and resolutions.

More recently, the focus of macro-design has shifted toward expanding the width of the residual stream~\citep{chai-etal-2020-highway, fang2023crosslayer, xie2023residualtransformerdualresidual, pagliardini2024denseformer, menghani2025laurel, heddes2025deepcrossattention, zhu2024hyper, mak2025residual, xiao2025muddformer}. Hyper-Connections (HC)~\citep{zhu2024hyper} introduced learnable matrices to modulate connection strengths among features at varying depths, while the Residual Matrix Transformer (RMT)~\citep{mak2025residual} replaced the standard residual stream with an outer-product memory matrix to facilitate feature storage. Similarly, MUDDFormer~\citep{xiao2025muddformer} employs multiway dynamic dense connections to optimize cross-layer information flow.
Despite their potential, these approaches compromise the inherent identity mapping property of the residual connection, thereby introducing instability and hindering scalability. Furthermore, they incur significant memory access overhead due to expanded feature widths.
Building upon HC, the proposed \mhcshort{} restricts the residual connection space onto a specific manifold to restore the identity mapping property, while also incorporating rigorous infrastructure optimizations to ensure efficiency. This approach enhances stability and scalability while maintaining the topological benefits of expanded connections.

\section{Preliminary}

We first establish the notation used in this work. In the HC formulation, the input to the $l$-th layer, $\textbf{x}_l\in \mathbb{R}^{1\times C}$, is expanded by a factor of $n$ to construct a hidden matrix $\textbf{x}_l = (\textbf{x}^\top_{l,0}, \ldots, \textbf{x}^\top_{l,n-1})^\top \in \mathbb{R}^{n \times C}$ which can be viewed as $n$-stream residual. This operation effectively broadens the width of the residual stream. To govern the read-out, write-in, and updating processes of this stream, HC introduces three learnable linear mappings—$\hpre{l}, \hpost{l}\in \mathbb{R}^{1\times n}$, and $\hres{l}\in \mathbb{R}^{n\times n}$. These mappings modify the standard residual connection shown in Eq.~\eqref{eqn:single_rc}, resulting in the formulation given in Eq.~\eqref{eqn:single_hc}.

In the HC formulation, learnable mappings are composed of two parts of coefficients: the input-dependent one and the global one, referred to as dynamic mappings and static mappings, respectively. Formally, HC computes the coefficients as follows:
\begin{equation}
    \begin{cases}
        \tilde{\mathbf{x}}_l = \text{RMSNorm}(\mathbf{x}_l) \\
        \hpre{l} = \alpha_l^\mathrm{pre} \cdot \tanh(\theta^\mathrm{pre}_l \tilde{\mathbf{x}}^\top_l) + \mathbf{b}_l^\mathrm{pre} \\
        \hpost{l} = \alpha_l^\mathrm{post} \cdot \tanh(\theta^\mathrm{post}_l \tilde{\mathbf{x}}^\top_l) + \mathbf{b}_l^\mathrm{post} \\
        \hres{l} = \alpha_l^\mathrm{res} \cdot \tanh(\theta^\mathrm{res}_l \tilde{\mathbf{x}}^\top_l) + \mathbf{b}_l^\mathrm{res}, \\
    \end{cases}
    \label{eqn:hc_details}
\end{equation}
where $\text{RMSNorm}(\cdot)$~\citep{zhang2019root} is applied to the last dimension, and the scalars $\alpha_l^\mathrm{pre}, \alpha_l^\mathrm{post}$ and $\alpha_l^\mathrm{res} \in \mathbb{R}$ are learnable gating factors initialized to small values. The dynamic mappings are derived via linear projections parameterized by $\theta^\mathrm{pre}_l, \theta^\mathrm{post}_l \in \mathbb{R}^{1 \times C}$ and $\theta^\mathrm{res}_l \in \mathbb{R}^{n \times C}$, while the static mappings are represented by learnable biases $\mathbf{b}_l^\mathrm{pre}, \mathbf{b}_l^\mathrm{post} \in \mathbb{R}^{1\times n}$ and $\mathbf{b}_l^\mathrm{res} \in \mathbb{R}^{n\times n}$.

It is worth noting that the introduction of these mappings—$\hpre{l}$, $\hpost{l}$, and $\hres{l}$—incurs negligible computational overhead, as the typical expansion rate $n$, e.g. 4, is much smaller than the input dimension $C$. With this design, HC effectively decouples the information capacity of the residual stream from the layer's input dimension, which is strongly correlated with the model's computational complexity (FLOPs). Consequently, HC offers a new avenue for scaling by adjusting the residual stream width, complementing the traditional scaling dimensions of model FLOPs and training data size discussed in pre-training scaling laws~\citep{Hoffmann2022Chinchilla}.

Although HC necessitates three mappings to manage the dimensional mismatch between the residual stream and the layer input, preliminary experiments presented in Tab.~\ref{tab:hc_ablation_preliminary} indicate that the residual mapping $\hres{l}$ yields the most significant performance gain. This finding underscores the critical importance of effective information exchange within the residual stream.

\begin{table}[htbp]
\centering
\caption{\textbf{Ablation Study of HC Components.} When a specific mapping ($\hpre{l}$, $\hpost{l}$, or $\hres{l}$) is disabled, we employ a fixed mapping to maintain dimensional consistency: uniform weights of $1/n$ for $\hpre{l}$, uniform weights of ones for $\hpost{l}$, and the identity matrix for $\hres{l}$.}
\label{tab:hc_ablation_preliminary}
\begin{tabular}{ccc|c}
  \toprule
  \hres{l} & \hpre{l} & \hpost{l}  & Absolute Loss Gap \\
  \midrule
   &   &   & $\phantom{-\ }0.0\phantom{00}$ \\
  \checkmark &   &   & $-\ 0.022$ \\
   \checkmark & \checkmark &   & $-\ 0.025$ \\
  \checkmark & \checkmark & \checkmark & $-\ 0.027$ \\
  \bottomrule
\end{tabular}
\end{table}

\subsection{Numerical Instability}
While the residual mapping $\hres{l}$ is instrumental for performance, its sequential application poses a significant risk to numerical stability. As detailed in Eq.~\eqref{eqn:multi_hc}, when HC is extended across multiple layers, the effective signal propagation from layer $l$ to $L$ is governed by the composite mapping $\prod_{i=1}^{L-l}\mathcal{H}_{L-i}^{\mathrm{res}}$.
Since the learnable mapping $\hres{l}$ is unconstrained, this composite mapping inevitably deviates from the identity mapping. Consequently, the signal magnitude is prone to explosion or vanishing during both the forward pass and backpropagation. This phenomenon undermines the fundamental premise of residual learning, which relies on unimpeded signal flow, thereby destabilizing the training process in deeper or larger-scale models.

\begin{figure}[h]
    \centering
    \includegraphics[width=1.0\textwidth]{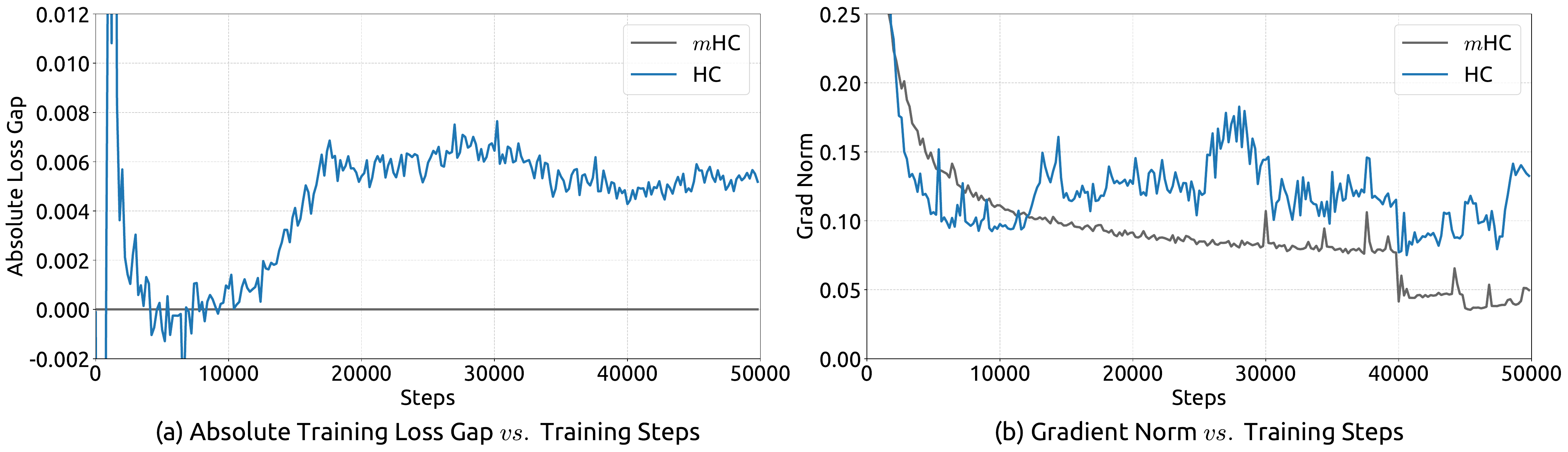}
    \caption{\textbf{Training Instability of Hyper-Connections (HC).} This figure illustrates (a) the absolute loss gap of HC relative to \mhcshort{}, and (b) the comparisons of gradient norms. All results are based on 27B models.}
    \label{fig:27b_loss_grad}
\end{figure}

\begin{figure}[h]
    \centering
    \includegraphics[width=1.0\textwidth]{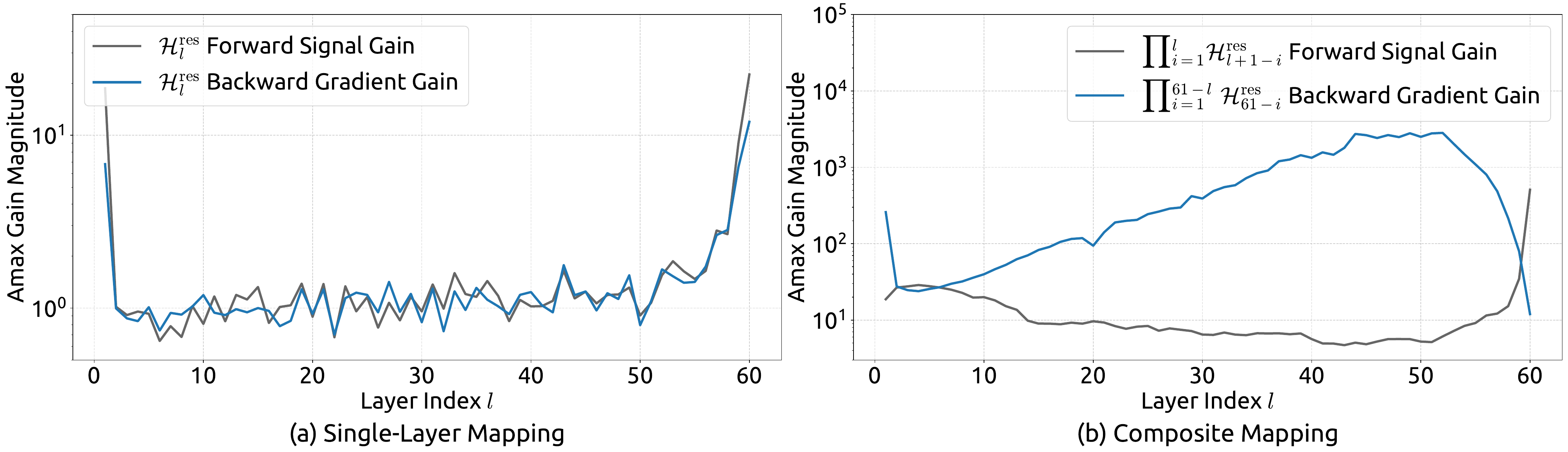}
    \caption{\textbf{Propagation Instability of Hyper-Connections (HC).} This figure illustrates the propagation dynamics of (a) the single-layer mapping $\mathcal{H}^{\mathrm{res}}_l$ and (b) the composite mapping $\prod_{i=1}^{L-l}\mathcal{H}_{L-i}^{\mathrm{res}}$ within the 27B model. The layer index $l$ (x-axis) unrolls each standard Transformer block into two independent layers (Attention and FFN). The Amax Gain Magnitude (y-axis) is calculated as the maximum absolute row sum (for the forward signal) and column sum (for the backward gradient), averaged over all tokens in a selected sequence.}
    \label{fig:27b_forward_backward_gain}
\end{figure}

Empirical evidence supports this analysis. We observe unstable loss behavior in large-scale experiments, as illustrated in Fig.~\ref{fig:27b_loss_grad}. Taking \mhcshort {} as the baseline, HC exhibits an unexpected loss surge around the 12k step, which is highly correlated with the instability in the gradient norm. Furthermore, the analysis on \hres{l} validates the mechanism of this instability.
To quantify how the composite mapping $\prod_{i=1}^{L-l}\mathcal{H}_{L-i}^{\mathrm{res}}$ amplifies signals along the residual stream, we utilize two metrics. The first, based on the maximum absolute value of the row sums of the composite mapping, captures the worst-case expansion in the forward pass. The second, based on the maximum absolute column sum, corresponds to the backward pass. We refer to these metrics as the \textit{Amax Gain Magnitude} of the composite mapping. As shown in Fig.~\ref{fig:27b_forward_backward_gain} (b), the Amax Gain Magnitude yields extreme values with peaks of 3000, a stark divergence from 1 that confirms the presence of exploding residual streams.

\subsection{System Overhead}

While the computational complexity of HC remains manageable due to the linearity of the additional mappings, the system-level overhead prevents a non-negligible challenge. Specifically, memory access (I/O) costs often constitute one of the primary bottlenecks in modern model architectures, which is widely referred to as the ``memory wall''~\citep{dao2022flashattention}. This bottleneck is frequently overlooked in architectural design, yet it decisively impacts runtime efficiency.

\begin{table}[h]
\centering
\caption{\textbf{Comparison of Memory Access Costs Per Token.} This analysis accounts for the overhead introduced by the residual stream maintenance in the forward pass, excluding the internal I/O of the layer function $\mathcal{F}$.}
\label{tab:memory_access}
\begin{tabular}{c|c c c}
\toprule
Method & Operation & Read (Elements) & Write (Elements) \\
\midrule
\multirow{2}{*}{\begin{tabular}[c]{@{}c@{}}Residual\\ Connection\end{tabular}} & Residual Merge & $2C$ & $C$ \\ \cmidrule{2-4}
 & \textbf{Total I/O} & $\mathbf{2C}$ & $\mathbf{C}$ \\
\midrule
\multirow{6}{*}{\begin{tabular}[c]{@{}c@{}}Hyper-\\ Connections\end{tabular}} & Calculate $\hpre{l}$, $\hpost{l}$, $\hres{l}$ & $nC$ & $n^2+2n$ \\
 & $\hpre{l}$ & $nC+n$ & $C$ \\
 & $\hpost{l}$ & $C+n$ & $nC$ \\
 & $\hres{l}$ & $nC+n^2$ & $nC$ \\
 & Residual Merge & $2nC$ & $nC$ \\ \cmidrule{2-4}
 & \textbf{Total I/O} & $\mathbf{(5n+1)C+n^2+2n}$ & $\mathbf{(3n+1)C+n^2+2n}$ \\
\bottomrule
\end{tabular}
\end{table}

Focusing on the widely adopted pre-norm Transformer~\citep{vaswani2017attention} architecture, we analyze the I/O patterns inherent to HC. Tab.~\ref{tab:memory_access} summarizes the per token memory access overhead in a single residual layer introduced by the $n$-stream residual design. The analysis reveals that HC increases the memory access cost by a factor approximately proportional to $n$. This excessive I/O demand significantly degrades training throughput without the mitigation of fused kernels. Besides, since $\hpre{l}$, $\hpost{l}$, and $\hres{l}$ involve learnable parameters, their intermediate activations are required for backpropagation. This results in a substantial increase in the GPU memory footprint, often necessitating gradient checkpointing to maintain feasible memory usage. Furthermore, HC requires $n$-fold more communication cost in pipeline parallelism~\citep{qi2024zero}, leading to larger bubbles and decreasing the training throughput.

\section{Method}
\subsection{\mhcfull{}}
Drawing inspiration from the identity mapping principle~\citep{he2016identity}, the core premise of \mhcshort{} is to constrain the residual mapping $\hres{l}$ onto a specific manifold.
While the original identity mapping ensures stability by enforcing $\hres{l} = \mathbf{I}$, it fundamentally precludes information exchange within the residual stream, which is critical for maximizing the potential of multi-stream architectures. Therefore, we propose projecting the residual mapping onto a manifold that simultaneously maintains the stability of signal propagation across layers and facilitates mutual interaction among residual streams to preserve the model's expressivity.
To this end, we restrict \hres{l} to be a doubly stochastic matrix, which has non-negative entries where both the rows and columns sum to 1. Formally, let $\mathcal{M}^\mathrm{res}$ denote the manifold of doubly stochastic matrices (also known as the Birkhoff polytope).
We constrain \hres{l} to $\mathcal{P}_{\mathcal{M}^\mathrm{res}}(\hres{l})$, defined as:
\begin{equation}
    \mathcal{P}_{\mathcal{M}^\mathrm{res}}(\hres{l}) \coloneq \left\{ \hres{l} \in \mathbb{R}^{n \times n} \mid \hres{l}\mathbf{1}_n = \mathbf{1}_n, \ \mathbf{1}^\top_n\hres{l} = \mathbf{1}^\top_n, \ \hres{l} \geq 0 \right\},
\end{equation}
where $\mathbf{1}_n$ represents the $n$-dimensional vector of all ones.

It is worth noting that when $n=1$, the doubly stochastic condition degenerates to the scalar $1$, thereby recovering the original identity mapping. The choice of double stochasticity confers several rigorous theoretical properties beneficial for large-scale model training:

\begin{enumerate}
    \item \textbf{Norm Preservation:} The spectral norm of a doubly stochastic matrix is bounded by 1 (i.e., $\|\hres{l}\|_2 \le 1$). This implies that the learnable mapping is non-expansive, effectively mitigating the gradient explosion problem.
    \item \textbf{Compositional Closure:} The set of doubly stochastic matrices is closed under matrix multiplication. This ensures that the composite residual mapping across multiple layers, $\prod_{i=1}^{L-l}\mathcal{H}_{L-i}^{\mathrm{res}}$, remains doubly stochastic, thereby preserving stability throughout the entire depth of the model.
    \item \textbf{Geometric Interpretation via the Birkhoff Polytope:} The set $\mathcal{M}^\mathrm{res}$ forms the Birkhoff polytope, which is the convex hull of the set of permutation matrices. This provides a clear geometric interpretation: the residual mapping acts as a convex combination of permutations. Mathematically, the repeated application of such matrices tends to increase the mixing of information across streams monotonically, effectively functioning as a robust feature fusion mechanism.
\end{enumerate}

Additionally, we impose non-negativity constraints on the input mappings $\hpre{l}$ and output mappings $\hpost{l}$. This constrain prevents signal cancellation arising from the composition of positive and negative coefficients, which can also be considered as a special manifold projection.

\subsection{Parameterization and Manifold Projection}

In this section, we detail the calculation process of $\hpre{l}, \hpost{l}, \text{and } \hres{l}$ in \mhcshort{}. Given the input hidden matrix $\mathbf{x}_l \in \mathbb{R}^{n \times C}$ at the $l$-th layer, we first flatten it into a vector $\vec{\mathbf{x}}_l = \text{vec}(\mathbf{x}_l) \in \mathbb{R}^{1\times nC}$ to preserve full context information. Then, we follow the original HC formulation to get the dynamic mappings and the static mappings as follows:
\begin{equation}
    \begin{cases}
        \vec{\mathbf{x}}'_l = \text{RMSNorm}(\vec{\mathbf{x}}_l) \\
        \tlhpre{l} = \alpha_l^\mathrm{pre} \cdot (\vec{\mathbf{x}}'_l\phi^\mathrm{pre}_l) + \mathbf{b}_l^\mathrm{pre} \\
        \tlhpost{l} = \alpha_l^\mathrm{post} \cdot (\vec{\mathbf{x}}'_l\phi^\mathrm{post}_l) + \mathbf{b}_l^\mathrm{post} \\
        \tlhres{l} = \alpha_l^\mathrm{res} \cdot \text{mat}(\vec{\mathbf{x}}'_l\phi^\mathrm{res}_l) + \mathbf{b}_l^\mathrm{res}, \\
    \end{cases}
\end{equation}
where $\phi^\mathrm{pre}_l, \phi^\mathrm{post}_l \in \mathbb{R}^{nC \times n}$ and $\phi^\mathrm{res}_l \in \mathbb{R}^{nC \times n^2}$ are linear projections for dynamic mappings and $\text{mat}(\cdot)$ is a reshape function from $\mathbb{R}^{1\times n^2}$ to $\mathbb{R}^{n\times n}$.

Then, the final constrained mappings are obtained via:
\begin{equation}
    \begin{cases}
        \hpre{l} = \sigma(\tlhpre{l}) \\
        \hpost{l} = 2\sigma(\tlhpost{l}) \\
        \hres{l} = \text{Sinkhorn-Knopp}(\tlhres{l}),
    \end{cases}
\end{equation}
where $\sigma(\cdot)$ denotes the Sigmoid function. The $\text{Sinkhorn-Knopp}(\cdot)$ operator firstly makes all elements to be positive via an exponent operator and then conducts iterative normalization process that alternately rescales rows and columns to sum to 1. Specifically, given a positive matrix $\mathbf{M}^{(0)} = \exp(\tlhres{l})$ as the start point, the normalization iteration proceeds as:
\begin{equation}
    \mathbf{M}^{(t)} = \mathcal{T}_r\left(\mathcal{T}_c(\mathbf{M}^{(t-1)})\right),
\end{equation}
where $\mathcal{T}_r$ and $\mathcal{T}_c$ denote row and column normalization, respectively. This process converges to a doubly stochastic matrix $\hres{l}=\mathbf{M}^{(t_{\text{max}})}$ as $t_{\text{max}} \to \infty$. We choose $t_{\text{max}}=20$ as a practical value in our experiments.

\subsection{Efficient Infrastructure Design}

In this section, we detail the infrastructure design tailored for \mhcshort{}.
Through rigorous optimization, we implement \mhcshort{} (with $n = 4$) in large-scale models with a marginal training overhead of only 6.7\%.
\subsubsection{Kernel Fusion}

Observing that RMSNorm in \mhcshort{} imposes significant latency when operating on the high-dimensional hidden state $\vec{\mathbf{x}}_l \in \mathbb{R}^{1\times nC}$, we reorder the dividing-by-norm operation to follow the matrix multiplication. This optimization maintains mathematical equivalence while improving efficiency.
Furthermore, we employ mixed-precision strategies to maximize numerical accuracy without compromising speed, and fuse multiple operations with shared memory access into unified compute kernels to reduce memory bandwidth bottlenecks.
Based on the inputs and parameters detailed in Eq.~\eqref{eq:fuse:1} to ~\eqref{eq:fuse:4}, we implement three specialized \mhcshort{} kernels to compute \hpre{l}, \hpost{l}, and \hres{l}. In these kernels, the biases and linear projections are consolidated into $\mathbf{b}_l$ and $\phi_l$, and the RMSNorm weight is also absorbed in $\phi_l$.

\begin{itemize}
    \item Eq.~\eqref{eq:fuse:5} to \eqref{eq:fuse:6}: We develop a unified kernel that fuses two scans on $\vec{\mathbf{x}}_l$, leveraging matrix multiplication units to maximize memory bandwidth utilization. The backward pass—comprising two matrix multiplications—is similarly consolidated into a single kernel, eliminating redundant reloading of $\vec{\mathbf{x}}_l$. Both kernels feature a finely tuned pipeline (load, cast, compute, store) to efficiently handle mixed-precision processing.
    \item Eq.~\eqref{eq:fuse:7} to ~\eqref{eq:fuse:9}: These lightweight operations on small coefficients are opportunistically fused into a single kernel, significantly reducing kernel launch overhead.
    \item Eq.~\eqref{eq:fuse:10}: We implement the Sinkhorn-Knopp iteration within a single kernel. For the backward pass, we derive a custom backward kernel that recomputes the intermediate results on-chip and traverses the entire iteration.
\end{itemize}

\vspace{-0.4in}

\begin{align}
    \phi_l                                                                      &: \text{tfloat32}          &&[nC, n^2+2n]                                                              \label{eq:fuse:1}\\
    \vec{\mathbf{x}}_l                                                                      &: \text{bfloat16}          &&[1, nC]                                                                      \label{eq:fuse:2}\\
    \alpha_l^\mathrm{pre}, \alpha_l^\mathrm{post}, \alpha_l^\mathrm{res}                                                    &: \text{float32}           &&\text{Scalars}                                                          \label{eq:fuse:3}\\
    \mathbf{b}_l                                                                      &: \text{float32}           &&[1, n^2+2n]                                                                  \label{eq:fuse:4}\\
    \left[{\ttlhpre{l}}, {\ttlhpost{l}}, {\ttlhres{l}}\right]   &: \text{float32}           &&= \vec{\mathbf{x}}_l\phi_l                                                  \label{eq:fuse:5}\\
    r                                                                               &: \text{float32}           &&= \left\|\vec{\mathbf{x}}_l\right\|_2 / \sqrt{nC}                                                       \label{eq:fuse:6}\\
    \left[\tlhpre{l}, \tlhpost{l}, \tlhres{l}\right]         &: \text{float32}           &&= 1/r \left[\alpha_l^\mathrm{pre}{\ttlhpre{l}}, \alpha_l^\mathrm{post}{\ttlhpost{l}}, \alpha_l^\mathrm{res}{\ttlhres{l}}\right] + \mathbf{b}_l \label{eq:fuse:7}\\
    \hpre{l}                                                                      &: \text{float32}           &&= \sigma\left(\tlhpre{l}\right)                                   \label{eq:fuse:8}\\
    \hpost{l}                                                                      &: \text{float32}           &&= 2\sigma\left(\tlhpost{l}\right)                                  \label{eq:fuse:9}\\
    \hres{l}                                                                      &: \text{float32}           &&= \text{Sinkhorn-Knopp}\left(\tlhres{l}\right)   \label{eq:fuse:10}
\end{align}

Using the coefficients derived from the aforementioned kernels, we introduce two additional kernels to apply these mappings: one for $\mathcal{F}_{\mathrm{pre}}\coloneq \hpre{l}\mathbf{x}_l$ and another for $\mathcal{F}_{\mathrm{post,res}}\coloneq \hres{l}\mathbf{x}_l+\mathcal{H}_{l}^{\mathrm{post}\, \top}\mathcal{F}(\cdot,\cdot)$. Through fusing the application of \hpost{l} and \hres{l} with residual merging, we reduce the number of elements read from $(3n+1)C$ to $(n+1)C$ and the number of elements written from $3nC$ to $nC$ for this kernel.
We efficiently implement the majority of kernels (excluding Eq.~\eqref{eq:fuse:5} to ~\eqref{eq:fuse:6}) using TileLang~\citep{wang2025tilelang}. This framework streamlines the implementation of kernels with complex calculation process and allows us to fully utilize the memory bandwidth with minimal engineering effort.
\subsubsection{Recomputing}

The $n$-stream residual design introduces substantial memory overhead during training.
To mitigate this, we discard the intermediate activations of the \mhcshort{} kernels after the forward pass and recompute them on-the-fly in the backward pass, through re-executing the \mhcshort{} kernels without the heavy layer function $\mathcal{F}$.
Consequently, for a block of $L_r$ consecutive layers, we need only store the input $\mathbf{x}_{l_0}$ to the first layer.
Excluding lightweight coefficients while accounting for the pre-norm with in $\mathcal{F}$, Tab.~ \ref{tab:actvation_memory} summarizes the intermediate activations preserved for the backward pass.

\begin{table}[h]
    \centering
    \caption{\textbf{Stored and Recomputed Intermediate Activations} We list per token activation preserved for the backward pass and the transient activation recomputed in $L_r$ consecutive layers. Layer $l_0$ represents the first layer in $L_r$ layers and layer $l$ is in $[l_0, l_0+L_r-1]$.}

    \begin{tabular}{c|c|c|c c c}
    \toprule
        Activations & $\mathbf{x}_{l_0}$ & $\mathcal{F}(\hpre{l}\mathbf{x}_l, \mathcal{W}_l)$ & $\mathbf{x}_{l}$ & $\hpre{l}\mathbf{x}_l$ & $\text{RMSNorm}(\hpre{l}\mathbf{x}_l)$  \\
    \midrule
        Size (Elements) & $nC$ & $C$ & $nC$ & $C$ & $C$ \\
        Stored Method & Every $L_r$ layers & Every layer & \multicolumn{3}{c}{Transient inside $L_r$ layers} \\
    \bottomrule
    \end{tabular}
    \label{tab:actvation_memory}
\end{table}

Since \mhcshort{} kernels recomputation is performed for blocks of $L_r$ consecutive layers, given a total of $L$ layers, we must persistently store the first layer input $\mathbf{x}_{l_0}$ for all $\lceil\tfrac{L}{L_r}\rceil$ blocks for the backward pass.
In addition to this resident memory, the recomputation process introduces a transient memory overhead of $(n+2)C\times L_r$ elements for the active block, which determines the peak memory usage during backpropagation. Consequently, we determine the optimal block size $L_r^*$ by minimizing the total memory footprint corresponded to $L_r$:
\begin{equation}
L_r^* = \arg\min_{L_r} \left[ nC\times \left\lceil\frac{L}{L_r}\right\rceil + (n+2)C\times L_r \right] \approx \sqrt{\frac{nL}{n+2}}.
\end{equation}

Furthermore, pipeline parallelism in large-scale training imposes a constraint: recomputation blocks must not cross pipeline stage boundaries. Observing that the theoretical optimum $L_r^*$ typically aligns with the number of layers per pipeline stage, we choose to synchronize the recomputation boundaries with the pipeline stages.

\subsubsection{Overlapping Communication in DualPipe}

\begin{figure}[h]
    \centering
    \includegraphics[width=\linewidth]{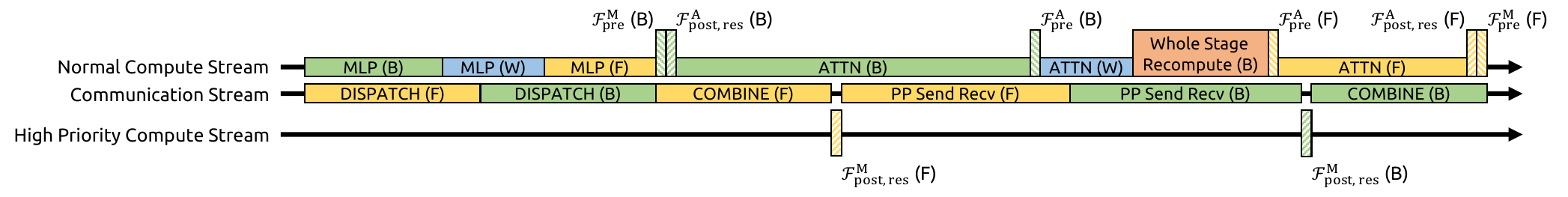}
    \caption{\textbf{Communication-Computation Overlapping for \mhcshort{}.}
    We extend the DualPipe schedule to handle the overhead introduced by \mhcshort{}. Lengths of each block are illustrative only and do not represent actual duration. (F), (B), (W) refers to forward pass, backward pass, weight gradient computation, respectively. $\mathcal{F}^{\mathrm{A}}\ \text{and}\ \mathcal{F}^{\mathrm{M}}$ represents kernels corresponded to Attention and MLP, respectively.}
    \label{fig:dualpipe}
\end{figure}

In large-scale training, pipeline parallelism is the standard practice for mitigating parameter and gradient memory footprints.
Specifically, we adopt the DualPipe schedule~\citep{liu2024deepseek_v3}, which effectively overlaps scale-out interconnected communication traffic, such as those in expert and pipeline parallelism.
However, compared to the single-stream design, the proposed $n$-stream residual in \mhcshort{} incurs substantial communication latency across pipeline stages.
Furthermore, at stage boundaries, the recomputation of \mhcshort{} kernels for all $L_r$ layers introduces non-negligible computational overhead.
To address these bottlenecks, we extend the DualPipe schedule (see Fig.~\ref{fig:dualpipe}) to facilitate improved overlapping of communication and computation at pipeline stage boundaries.

Notably, to prevent blocking the communication stream, we execute the $\mathcal{F}_{\mathrm{post,res}}$ kernels of MLP (i.e. FFN) layers on a dedicated high-priority compute stream. We further refrain from employing persistent kernels for long-running operations in attention layers, thereby preventing extended stalls. This design enables the preemption of overlapped attention computations, allowing for flexible scheduling while maintaining high utilization of the compute device's processing units.
Furthermore, the recomputation process is decoupled from pipeline communication dependencies, as the initial activation of each stage $\mathbf{x}_{l_0}$ is already cached locally.

\section{Experiments}

\subsection{Experimental Setup}

We validate the proposed method via language model pre-training, conducting a comparative analysis between the baseline, HC, and our proposed \mhcshort{}.
Utilizing MoE architectures inspired by DeepSeek-V3~\citep{liu2024deepseek_v3}, we train four distinct model variants to cover different evaluation regimes.
Specifically, the expansion rate $n$ for both HC and \mhcshort{} is set to 4.
Our primary focus is a 27B model trained with a dataset size proportional to its parameters, which serves as the subject for our system-level main results.
Expanding on this, we analyze the compute scaling behavior by incorporating smaller 3B and 9B models trained with proportional data, which allows us to observe performance trends across varying compute.
Additionally, to specifically investigate the token scaling behavior, we train a separate 3B model on a fixed corpus of 1 trillion tokens.
Detailed model configurations and training hyper-parameters are provided in Appendix~\ref{appendix:model_specs}.

\subsection{Main Results}

\begin{figure}[h]
    \centering
    \includegraphics[width=1.0\textwidth]{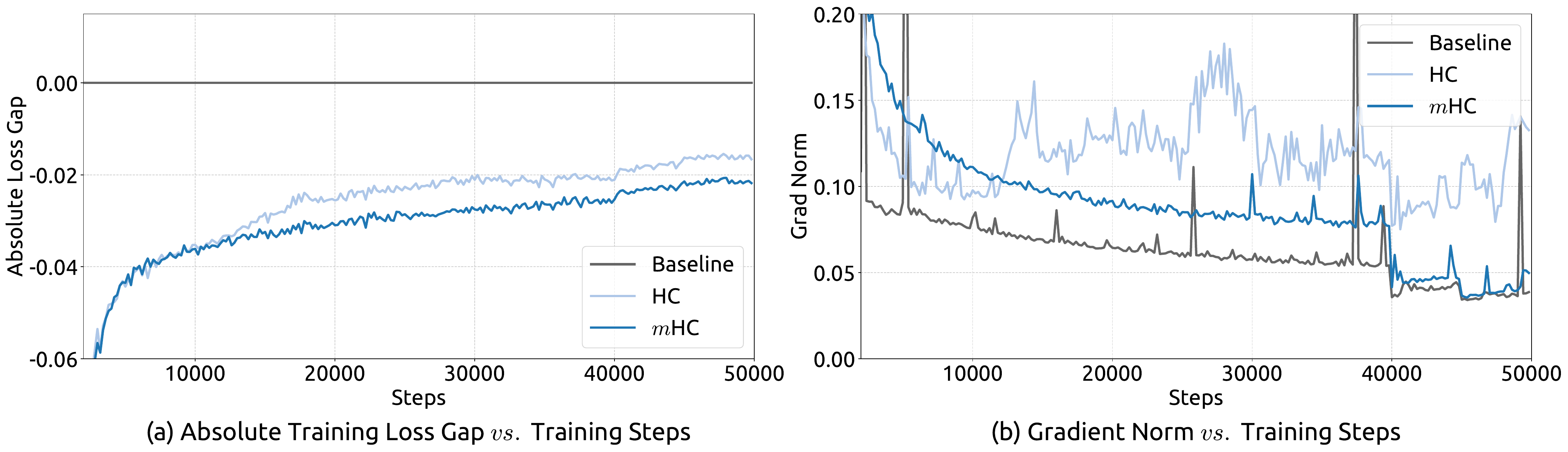}
    \caption{
    \textbf{Training Stability of \mhcfull{} (\mhcshort{}).} This figure illustrates (a) the absolute training loss gap of \mhcshort{} and HC relative to the baseline, and (b) the gradient norm of the three methods. All experiments utilize the 27B model. The results demonstrate that \mhcshort{} exhibits improved stability in terms of both loss and gradient norm.
    }
    \label{fig:27b_loss_grad_all}
\end{figure}

\begin{table}[h]
    \centering
    \caption{
    \textbf{System-level Benchmark Results for 27B Models.} This table compares the zero-shot and few-shot performance of the Baseline, HC, and \mhcshort{} across 8 diverse downstream benchmarks. \mhcshort{} consistently outperforms the Baseline and surpasses HC on the majority of benchmarks, demonstrating its effectiveness in large-scale pre-training.
    }
    \footnotesize
    \setlength{\tabcolsep}{5pt}
    \begin{tabularx}{\linewidth}{l | Y Y Y Y Y Y Y Y}
    \toprule
    \textbf{Benchmark} & BBH & DROP & GSM8K & HellaSwag & MATH & MMLU & PIQA & TriviaQA \\
    \textbf{(Metric)} & (EM) & (F1) & (EM) & (Acc.) & (EM) & (Acc.) & (Acc.) & (EM) \\
    \midrule
    \textbf{\# Shots} & 3-shot & 3-shot & 8-shot & 10-shot & 4-shot & 5-shot & 0-shot & 5-shot \\
    \midrule
    \textbf{27B Baseline}
    & 43.8 & 47.0 & 46.7 & 73.7 & 22.0 & 59.0 & 78.5 & 54.3 \\
    \textbf{27B w/ HC}
    & 48.9 & 51.6 & 53.2 & 74.3 & \textbf{26.4} & 63.0 & 79.9 & 56.3 \\
    \rowcolor[HTML]{E6F3FD} \textbf{27B w/ \mhcshort{}}
    & \textbf{51.0} & \textbf{53.9} & \textbf{53.8} & \textbf{74.7} & 26.0 & \textbf{63.4} & \textbf{80.5} & \textbf{57.6} \\
    \bottomrule
    \end{tabularx}
    \label{tab:27b_result}
\end{table}

We begin by examining the training stability and convergence of the 27B models. As illustrated in Fig.~\ref{fig:27b_loss_grad_all} (a), \mhcshort{ } effectively mitigates the training instability observed in HC, achieving a final loss reduction of 0.021 compared to the baseline. This improved stability is further corroborated by the gradient norm analysis in Fig.~\ref{fig:27b_loss_grad_all} (b), where \mhcshort{} exhibits significantly better behavior than HC, maintaining a stable profile comparable to the baseline.

Tab.~\ref{tab:27b_result} presents the downstream performance across a diverse set of benchmarks~\citep{mmlu, gsm8k, hellaswag, hendrycks2021measuring, piqa, joshi-etal-2017-triviaqa}. \mhcshort{} yields comprehensive improvements, consistently outperforming the baseline and surpassing HC on the majority of tasks. Notably, compared to HC, \mhcshort{} further enhances the model's reasoning capabilities, delivering performance gains of 2.1\% on BBH~\citep{bbh} and 2.3\% on DROP~\citep{drop}.

\subsection{Scaling Experiments}

\begin{figure}[h]
    \centering
    \includegraphics[width=1.0\textwidth]{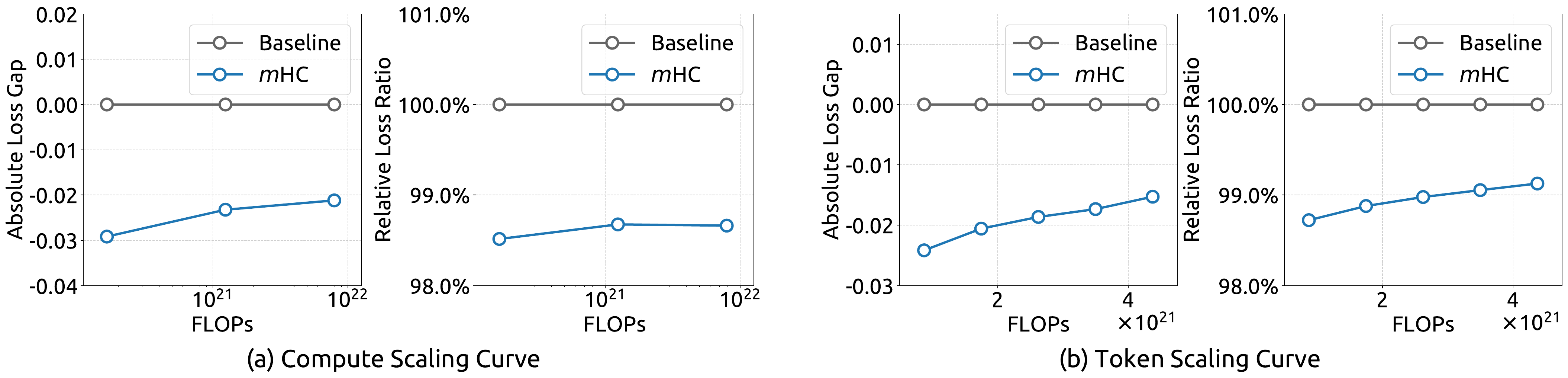}
    \caption{
    \textbf{Scaling properties of \mhcshort{} compared to the Baseline.} 
    \textbf{(a) Compute Scaling Curve.} 
    Solid lines depict the performance gap across different compute budgets. Each point represents a specific compute-optimal configuration of model size and dataset size, scaling from 3B and 9B to 27B parameters. 
    \textbf{(b) Token Scaling Curve.} 
    Trajectory of the 3B model during training. Each point represents the model's performance at different training tokens. 
    Detailed architectures and training configurations are provided in Appendix~\ref{appendix:model_specs}.
    }
    \label{fig:scaling_curve}
\end{figure}

To assess the scalability of our approach, we report the relative loss improvement of \mhcshort{} against the baseline across different scales.
In Fig.~\ref{fig:scaling_curve} (a), we plot the compute scaling curve spanning 3B, 9B, and 27B parameters.
The trajectory indicates that the performance advantage is robustly maintained even at higher computational budgets, showing only marginal attenuation.
Furthermore, we examine the within-run dynamics in Fig.~\ref{fig:scaling_curve} (b), which presents the token scaling curve for the 3B model.
Collectively, these findings validate the effectiveness of \mhcshort{} in large-scale scenarios. This conclusion is further corroborated by our in-house large-scale training experiments.

\subsection{Stability Analysis}

\begin{figure}[h]
    \centering
    \includegraphics[width=1.0\textwidth]{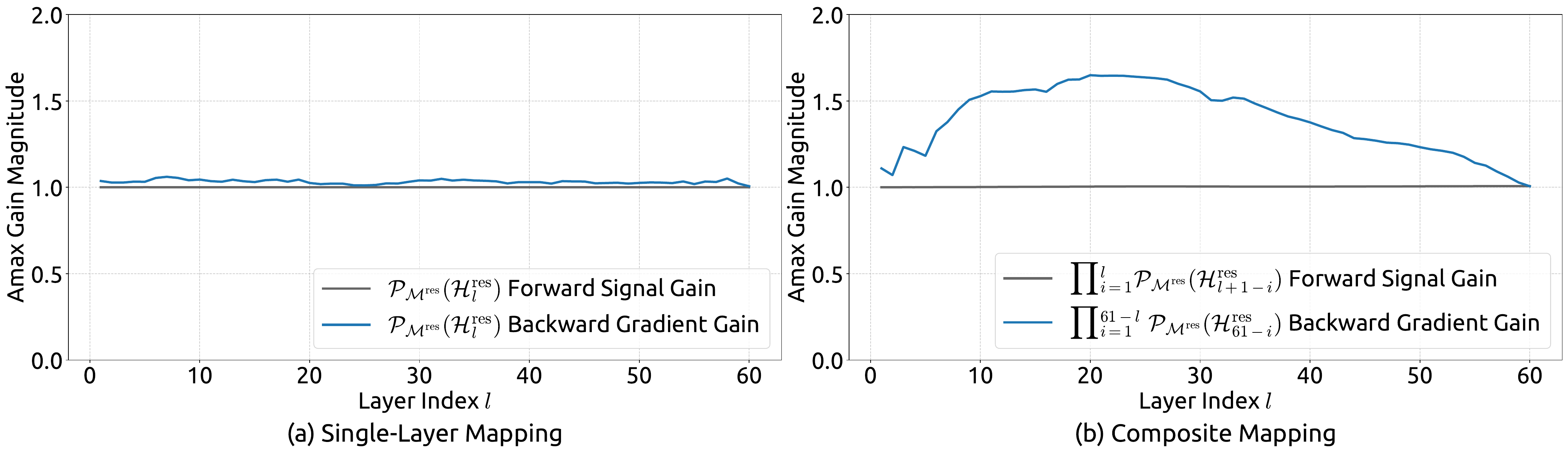}
    \caption{
    \textbf{Propagation Stability of \mhcfull{} (\mhcshort{}).} This figure illustrates the propagation dynamics of (a) the single-layer mapping $\mathcal{P}_{\mathcal{M}^{\mathrm{res}}}(\mathcal{H}^{\mathrm{res}}_l)$ and (b) the composite mapping $\prod_{i=1}^{L-l}\mathcal{P}_{\mathcal{M}^{\mathrm{res}}}(\mathcal{H}_{L-i}^{\mathrm{res}})$ within the 27B model. The results demonstrate that \mhcshort{} significantly enhances propagation stability compared to HC.
    }
    \label{fig:27b_mhc_forward_backward_gain}
\end{figure}

\begin{figure}[h]
    \centering
    \includegraphics[width=1.0\textwidth]{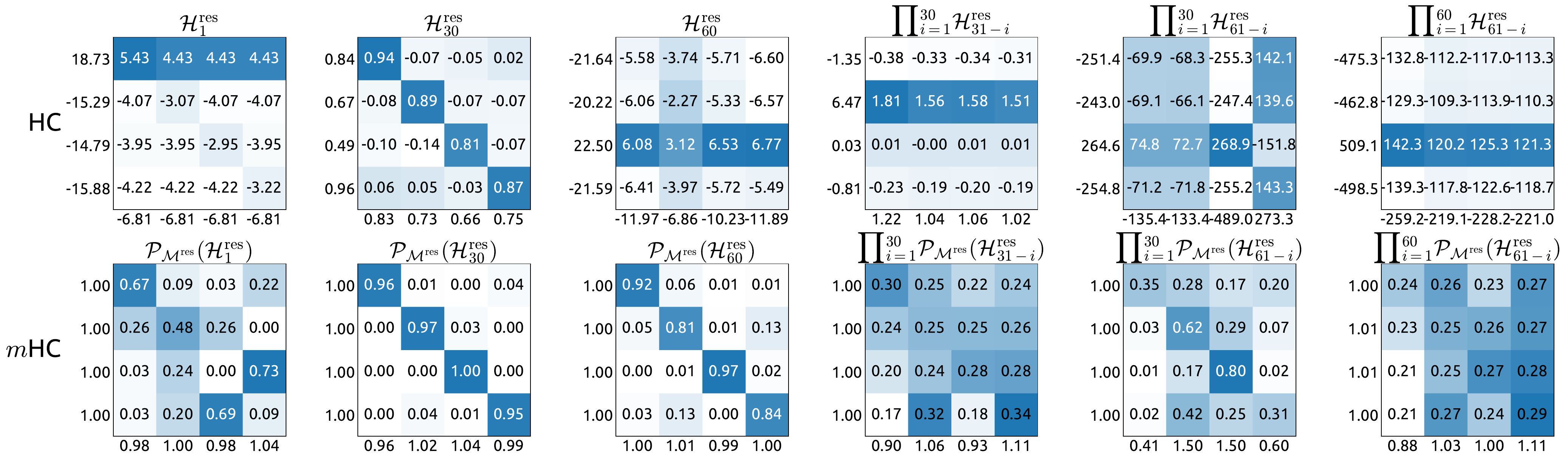}
    \caption{
    \textbf{Visualizations of Learnable Mappings.} This figure displays representative single-layer and composite mappings for HC (first row) and \mhcshort{} (second row). Each matrix is computed by averaging over all tokens within a selected sequence. The labels annotated along the y-axis and x-axis indicate the forward signal gain (row sum) and the backward gradient gain (column sum), respectively.
    }
    \label{fig:27b_hc_mhc_heatmap}
\end{figure}

Similar to Fig.~\ref{fig:27b_forward_backward_gain}, Fig.~\ref{fig:27b_mhc_forward_backward_gain} illustrates the propagation stability of \mhcshort{}. Ideally, the single-layer mapping satisfies the doubly stochastic constraint, implying that both the forward signal gain and the backward gradient gain should equal to 1. However, practice implementations utilizing the Sinkhorn-Knopp algorithm must limit the number of iterations to achieve computational efficiency. In our settings, we use 20 iterations to obtain an approximate solution. Consequently, as shown in Fig.~\ref{fig:27b_mhc_forward_backward_gain}(a), the backward gradient gain deviates slightly from 1. In the composite case shown in Fig.~\ref{fig:27b_mhc_forward_backward_gain}(b), the deviation increases but remains bounded, reaching a maximum value of approximately 1.6.
Notably, compared to the maximum gain magnitude of nearly 3000 in HC, \mhcshort{} significantly reduces it by three orders of magnitude.
These results demonstrate that \mhcshort{} significantly enhances propagation stability compared to HC, ensuring stable forward signal and backward gradient flows.
Additionally, Fig.~\ref{fig:27b_hc_mhc_heatmap} displays representative mappings. We observe that for HC, when the maximum gain is large, other values also tend to be significant, which indicates general instability across all propagation paths. In contrast, \mhcshort{} consistently yields stable results.

\section{Conclusion and Outlook}

In this paper, we identify that while expanding the width of residual stream and diversifying connections yields performance gains as proposed in Hyper-Connections (HC), the unconstrained nature of these connections leads to signal divergence.
This disruption compromises the conservation of signal energy across layers, inducing training instability and hindering the scalability of deep networks.
To address these challenges, we introduce \mhcfullbf{} (\mhcshortbf{}), a generalized framework that projects the residual connection space onto a specific manifold.
By employing the Sinkhorn-Knopp algorithm to enforce a doubly stochastic constraint on residual mappings, \mhcshort{} transforms signal propagation into a convex combination of features.
Empirical results confirm that \mhcshort{} effectively restores the identity mapping property, enabling stable large-scale training with superior scalability compared to conventional HC.
Crucially, through efficient infrastructure-level optimizations, \mhcshort{} delivers these improvements with negligible computational overhead.

As a generalized extension of the HC paradigm, \mhcshort{} opens several promising avenues for future research.
Although this work utilizes doubly stochastic matrices to ensure stability, the framework accommodates the exploration of diverse manifold constraints tailored to specific learning objectives.
We anticipate that further investigation into distinct geometric constraints could yield novel methods that better optimize the trade-off between plasticity and stability.
Furthermore, we hope \mhcshort{} rejuvenates community interest in macro-architecture design. By deepening the understanding of how topological structures influence optimization and representation learning, \mhcshort{} will help address current limitations and potentially illuminate new pathways for the evolution of next-generation foundational architectures.

\bibliography{main}

\newpage
\appendix

\section{Appendix}

\vspace{-0.075in}

\subsection{Detailed Model Specifications and Hyper-parameters.}
\label{appendix:model_specs}

\vspace{-0.075in}

\begin{table}[h!]
    \centering
    \caption{
    \textbf{Detailed Model Specifications and Hyper-parameters.} This table presents the architectural configurations for the 3B, 9B, and 27B models based on the DeepSeek-V3~\citep{liu2024deepseek_v3} architecture. It outlines the specific hyper-parameters for \mhcshort{} and HC, including the residual stream expansion and Sinkhorn-Knopp settings, alongside the optimization and training protocols used in the experiments.
    }
    \vspace{-0.05in}
    \begin{tabularx}{\linewidth}{l | Y Y Y | Y}
    \toprule
    \multirow{2}{*}{\textbf{Attribute}} & \multirow{2}{*}{3B} & \multirow{2}{*}{9B} & \multirow{2}{*}{27B} & 3B \\
    & & & & 1T Tokens \\
    \midrule
    Vocab Params & 331M & 496M & 662M & 331M \\
    Active Params & 612M & 1.66B & 4.14B & 612M \\
    Total Params & 2.97B & 9.18B & 27.0B & 2.97B \\
    \midrule
    Layers & 12 & 18 & 30 & 12 \\
    Leading Dense Layers & \multicolumn{3}{c|}{1} & 1\\
    Routed Experts & 64 & 64 & 72 & 64 \\
    Active Experts & \multicolumn{3}{c|}{6} & 6 \\
    Shared Experts & \multicolumn{3}{c|}{2} & 2 \\
    Dimension & 1280 & 1920 & 2560 & 1280 \\
    FFN Dimension & 896 & 1280 & 1536 & 896 \\
    Load Balancing Method & \multicolumn{3}{c|}{Loss-Free~\citep{wang2024auxiliary}} & Loss-Free \\
    Attention Heads & 16 & 24 & 32 & 16 \\
    Attention Dimension & \multicolumn{3}{c|}{128} & 128 \\
    Attention Variant & \multicolumn{3}{c|}{MLA~\citep{liu2024deepseek}} & MLA \\
    KV Rank & \multicolumn{3}{c|}{512} & 512 \\
    Position Embedding & \multicolumn{3}{c|}{RoPE~\citep{su2024roformer}} & RoPE \\
    RoPE Dimension & \multicolumn{3}{c|}{64} & 64 \\
    RoPE $\theta$ & \multicolumn{3}{c|}{10000} & 10000 \\
    Layer Norm Type & \multicolumn{3}{c|}{RMSNorm~\citep{zhang2019root}} & RMSNorm \\
    Layer Norm $\epsilon$ & \multicolumn{3}{c|}{1e-20} & 1e-20 \\
    \midrule
    \mhcshort{}/HC Expansion Rate $n$ & \multicolumn{3}{c|}{4} & 4 \\
    \mhcshort{}/HC Gating Factor Init $\alpha$ & \multicolumn{3}{c|}{0.01} & 0.01 \\
    \mhcshort{} Sinkhorn-Knopp $t_{\text{max}}$ & \multicolumn{3}{c|}{20} & 20 \\
    \midrule
    Sequence Length & \multicolumn{3}{c|}{4096} & 4096 \\
    Vocab Size & \multicolumn{3}{c|}{129280} & 129280 \\
    Batch Size & 320 & 512 & 1280 & 2560 \\
    Training Steps & 30000 & 50000 & 50000 & 100000 \\
    Training Tokens & 39.3B & 105B & 262B & 1.05T \\
    Warmup Steps & \multicolumn{3}{c|}{2000} & 2000 \\
    Optimizer & \multicolumn{3}{c|}{AdamW~\citep{loshchilov2017decoupled}} & AdamW \\
    AdamW Betas & \multicolumn{3}{c|}{(0.9, 0.95)} & (0.9, 0.95) \\
    AdamW $\epsilon$ & \multicolumn{3}{c|}{1e-20} & 1e-20 \\
    Base Learning Rate & 8.6e-4 & 5.9e-4 & 4.0e-4 & 9.0e-4 \\
    Lr Scheduler & \multicolumn{3}{c|}{Step} & Step\\
    Lr Decay Step Ratio & \multicolumn{3}{c|}{[0.8 $\times$, 0.9 $\times$]} & [0.8 $\times$, 0.9 $\times$] \\
    Lr Decay Rate & \multicolumn{3}{c|}{[0.316, 0.1]} & [0.316, 0.1] \\
    Weight Decay & \multicolumn{3}{c|}{0.1} & 0.1 \\
    \bottomrule
    \end{tabularx}
    \label{tab:model_specs}
    \vspace{-0.3in}
\end{table}

\setcounter{figure}{0}
\makeatletter 
\renewcommand{\thefigure}{A\@arabic\c@figure}
\makeatother

\setcounter{table}{0}
\makeatletter 
\renewcommand{\thetable}{A\@arabic\c@table}
\makeatother

\end{CJK*}
\end{document}